\documentclass[10pt,twocolumn,letterpaper]{article}

\usepackage[final,algorithms]{wacv}      
\usepackage{wacv}              

\usepackage{graphicx}
\usepackage{amsmath}

\usepackage{amssymb}
\usepackage{booktabs}
\usepackage[ruled]{algorithm2e}
\usepackage{multirow}
\usepackage[accsupp]{axessibility} 

\usepackage[pagebackref,breaklinks,colorlinks]{hyperref}

\usepackage[capitalize]{cleveref}
\crefname{section}{Sec.}{Secs.}
\Crefname{section}{Sec.}{Secs.}
\Crefname{table}{Table}{Tables}
\crefname{table}{Table}{Tables}

\def\wacvPaperID{2463} 
\def\confName{WACV}
\def\confYear{2025}
\newcommand{\cfg}{\mathrm{CFG}}
\newcommand{\Pt}{P_t}

\begin{document}

\title{When Does High-CFG Diffusion Inversion Fail?
A Controlled Study of Prompt--Latent Interactions
}



\author{Yan Zeng$^1$
~~~~Yusuke Hosoya$^{1}$
~~~~Huyen T. T. Tran$^{1}$
~~~~Takayuki Okatani$^{1,2}$\\
$^1$Graduate School of Information Sciences, Tohoku University ~~~~ $^2$RIKEN Center for AIP \\
}

\maketitle

\begin{abstract}
Text-guided diffusion inversion is central to image editing, where
an image is mapped to an initial latent and then edited by replaying the
denoising process under a modified prompt. In practice, however,
inversion is often performed with a lower classifier-free guidance
(CFG) scale than the one used for generation or editing. This mismatch
is empirically useful but leaves a basic question unresolved: when a
target image is generated by a high-CFG trajectory, when can that
trajectory actually be inverted?
We study this question in a controlled
generation--inversion--reconstruction setting, where the true initial
latent and denoising trajectory are known. Using prompts taken from an
existing diffusion-editing benchmark, we generate images under high CFG
and reconstruct them with fixed-point inversion using the same prompt
and guidance setting. 
The results reveal three types of prompt-level reconstruction behavior:
easy prompts that reconstruct for most initial latents, hard prompts
that fail for most initial latents, and intermediate prompts whose
success depends on the prompt--latent pairing. To analyze the
generation side, we define prompt pressure, a step-wise measure of how
strongly CFG moves the denoising update away from the unconditional
trajectory. Total pressure correlates with reconstruction quality and
separates easy from hard prompts, but it does not explain the success or
failure of intermediate prompt--latent pairs. Text-side analyses further
show that the main visual subject and wording can change inversion
difficulty. Finally, we evaluate a compact trajectory-consistency
intervention that relaxes guidance only at locally unstable inverse
steps. This diagnostic check improves reconstruction and
Prompt-to-Prompt editing in our controlled setting, supporting the view
that high-CFG inversion failure requires local, trajectory-aware
analysis.
\end{abstract}

\section{Related Work}
\label{sec:related}
Diffusion inversion methods usually aim to recover a noise code, latent trajectory, or optimized conditioning state that can reconstruct an image under a diffusion sampler. DDIM inversion follows the deterministic DDIM trajectory~\cite{song2021ddim}, while editing-oriented methods such as Null-text inversion~\cite{mokady2023nulltext}, Prompt-to-Prompt~\cite{hertz2022prompt}, and AIDI~\cite{pan2023aidi} improve reconstruction or editability by changing the inverse optimization or the conditioning used during editing. Recent fixed-point analyses provide theoretical support for viewing DDIM inversion as an implicit fixed-point problem~\cite{hang2024fixedpoint, meiri2023fixed}, and exact-inversion alternatives replace or modify the reverse dynamics through coupled transformations or solver-specific inverse updates~\cite{wallace2023edict,hong2024exact}. Other work improves inversion by changing the diffusion noise schedule~\cite{lin2024scheduleyouredit}, by distilling invertible few-step models with dynamic guidance~\cite{starodubcev2024icd}, or by introducing adaptive or reversible inversion rules~\cite{chen2025polaris,dai2024erddci}.

Classifier-free guidance~\cite{ho2022cfg} is central to these pipelines because it improves text alignment, but it also amplifies the prompt-dependent part of the denoising update. Several recent studies show that fixed or large guidance scales can distort the sampling geometry. CFG++ attributes DDIM inversion difficulty and mode collapse to off-manifold behavior under standard CFG~\cite{chung2025cfgpp}, while invertible consistency distillation observes that high CFG exacerbates reconstruction errors and that dynamic guidance can reduce them~\cite{starodubcev2024icd}. Feedback Guidance similarly treats guidance as trajectory- and state-dependent rather than as a fixed global hyperparameter~\cite{koulischer2025feedback}. These works motivate our focus on high-$\cfg$ inversion, but our goal is diagnostic: we analyze which prompts and prompt-seed trajectories enter unstable inversion regimes.

Our analysis is also related to generation-side studies of temporal
attention, initial noise, and seed reliability. T-GATE shows that text-conditioned cross-attention is especially important early in generation, suggesting that prompt influence is temporally structured \cite{liu2025tgate}. InitNO argues that not all initial noises are equally valid for a given prompt \cite{guo2024initno}, and reliable-seed analysis shows that some seeds systematically improve compositional text-to-image generation \cite{li2025reliableseeds}. Attention Refocusing studies failures in multi-object, attribute, and spatial prompts through attention behavior \cite{phung2024attentionrefocusing}. Finally, analyses of DDIM-inverted latents show that recovered noise can contain image-dependent structure rather than behaving like clean Gaussian noise \cite{staniszewski2026there}. Together, these works suggest that inversion failure should be studied
jointly over prompt content, initial latents, timesteps,
and guidance dynamics.

This paper is complementary to that line of work. Rather than proposing
a new general-purpose inversion framework, we ask which generated
samples become hard to invert and why. Our focus is the interaction
between prompt content, initial latents, and high-CFG
trajectory dynamics. The compact intervention in
Section~\ref{sec:diagnostic_intervention} is used only as a diagnostic
check of the trajectory-consistency implication.
\section{Problem Setup}
\label{sec:setup}




\subsection{Notation and inversion protocol}
\label{sec:notation}

We consider a latent diffusion model (LDM) with a VAE encoder
\(E_{\mathrm{VAE}}\) and decoder \(D_{\mathrm{VAE}}\) \cite{rombach2022high}. Let \(p\) denote a
text prompt and let \(z \equiv z_T\) denote the initial Gaussian latent.
For a fixed classifier-free guidance scale \(\omega\), the deterministic
DDIM sampler produces a latent denoising trajectory
\(z_T \rightarrow z_{T-1} \rightarrow \cdots \rightarrow z_0\), with
\(z_{t-1} = \phi_t(z_t; p,\omega)\). We denote the full denoising process by
\(F(p,z;\omega)=z_0\), and the corresponding generated RGB image by
\begin{equation}
G(p,z;\omega) = D_{\mathrm{VAE}}(F(p,z;\omega)).
\end{equation}

In implementation, the initial latent is sampled by a pseudorandom
Gaussian generator. For an integer seed \(s\), let
\begin{equation}
    \mathcal{Z}_s = \Gamma(s)
    = \{z_{p,s}\}
\end{equation}
be the corresponding list of initial latents, where \(\Gamma\) denotes the fixed
sampling routine. The seed \(s\) itself has no direct role in the
denoising dynamics; it only indexes the reproducible latent list \(\mathcal{Z}_s\).
We write the generated image as
\begin{equation}
x_{p,s} := G(p,z_{p,s};\omega).
\end{equation}
For inversion analysis, we use the terminal latent \(z_{0,p,s}\)
directly from the generation trajectory as the input to inversion. This avoids the additional error introduced by decoding \(z_{0,p,s}\) to pixel
space and then re-encoding it through the VAE. The recovered initial latent
is
\begin{equation}
\hat{z}_{p,s} = I(p,z_{0,p,s};\omega),
\end{equation}
and the reconstruction is obtained by replaying the same denoising
process:
\begin{equation}
\hat{x}_{p,s}
=
G(p,\hat{z}_{p,s};\omega)
=
D_{\mathrm{VAE}}(F(p,\hat{z}_{p,s};\omega)).
\end{equation}

\subsection{Experimental setup}
\label{sec:experimental_setup}

For the diagnostic analysis, we instantiate \(I\) with fixed-point
inversion (FPI). DDIM inversion is a common baseline, but its local
linearization error can make failures attributable to the inverse
approximation itself. FPI instead solves each inverse DDIM step as an
implicit fixed-point problem, giving a high-fidelity, lightweight
baseline without per-image embedding optimization or auxiliary coupled
trajectories~\cite{meiri2023fixed}. Recent strong iterative inversion methods, including AIDI, also rely on
iterative refinement or stabilization of the inverse trajectory
~\cite{pan2023aidi, parmar2023zero, garibi2024renoise}.
Thus, plain FPI serves as a clean core baseline for
diagnosing high-CFG trajectory difficulty.

We use the Stable Diffusion v1.4 checkpoint from CompVis at
\(512\times512\) resolution. Generation, inversion, and reconstruction
use the DDIM sampler with 50 steps and the same classifier-free
guidance scale, \(\omega=7\), unless a later ablation explicitly states
otherwise. The baseline FPI solver uses the deterministic fixed-point
update without relaxation, i.e., the new iterate is the raw one-step
inverse update. At each inverse timestep, the fixed-point residual is
the latent-space mean squared error between the current iterate and the
updated iterate. The solver stops when the residual increases relative
to the previous iteration. No separate hard maximum on the number of
FPI iterations is imposed in the baseline runs, since the iteration
typically either converges very quickly or diverges rapidly.

We evaluate inversion quality by the PSNR between the generated image
\(x_{p,s}\) and the reconstruction \(\hat{x}_{p,s}\). Reconstruction
PSNR is computed after VAE decoding, in RGB image space, using pixel
values on the 0--255 scale. CLIP image similarity and CLIP text
scores~\cite{radford2021learning} are computed with the OpenAI CLIP
ViT-B/32 backbone.

We use the 700 prompts from the \texttt{original\_prompt} field of
PIE-Bench~\cite{ju2024pnp}. In Section~\ref{sec:prompt_seed},  each prompt is evaluated with the
same ten seed-indexed initial latents \(\{z_s\}_{s=1}^{10}\), so that
\(z_{p,s}=z_s\) for all prompts \(p\). This shared-latent setting allows us
to separate the effects of prompt and initial latent on reconstruction
quality. In later sections, where this controlled comparison is not
needed, we sample initial latents independently for each prompt and seed
to avoid restricting image diversity.

The benchmark contains a balanced \(700 \times 10\) grid, giving 7000
prompt--seed pairs. Each prompt is evaluated with ten seed-indexed
initial latents, and each seed-indexed latent is reused across all
prompts. Table~\ref{tab:aggregate_psnr} reports the aggregate PSNR
statistics used in this and the next section. The first row summarizes
all 7000 prompt--seed pairs: the mean PSNR is 40.61 dB with a standard
deviation of 20.35 dB, and the values range from 6.08 dB to 69.61 dB.
This broad distribution indicates that FPI is sometimes nearly exact
and sometimes fails severely, making the benchmark suitable for factor
analysis.

\section{Variation across prompts and initial latents}
\label{sec:prompt_seed}

We first examine how FPI reconstruction quality varies across text
prompts and seed-indexed initial latents. These two factors enter the
generation process in different ways: the prompt specifies the
text-conditioned denoising dynamics, while the initial latent \(z_s\)
specifies the starting point of the trajectory. For readability, we use \(s\) and \(z_s\), \(S\) and \(\{z_s\}\) interchangeably in this sections. Since our benchmark
forms a complete \(700\times10\) prompt--seed grid, we can summarize
reconstruction quality along either axis and ask what kind of variation
is visible at the prompt level and at the latent level.

\subsection{Reconstruction behavior across initial latents}
\label{sec:prompt_regimes}

For each prompt \(p\), we summarize its reconstruction behavior across
the ten seed-indexed initial latents by the mean and standard deviation
of PSNR:
\[
\mu_p = \frac{1}{|S|}\sum_{s\in S}\mathrm{PSNR}(p,s),
\qquad
\sigma_p = \mathrm{Std}_{s\in S}[\mathrm{PSNR}(p,s)].
\]
The mean \(\mu_p\) measures the average reconstruction quality for the
prompt, while \(\sigma_p\) measures how much that quality changes across
initial latents.

\begin{figure}[t]
\centering
\includegraphics[width=0.95\linewidth]{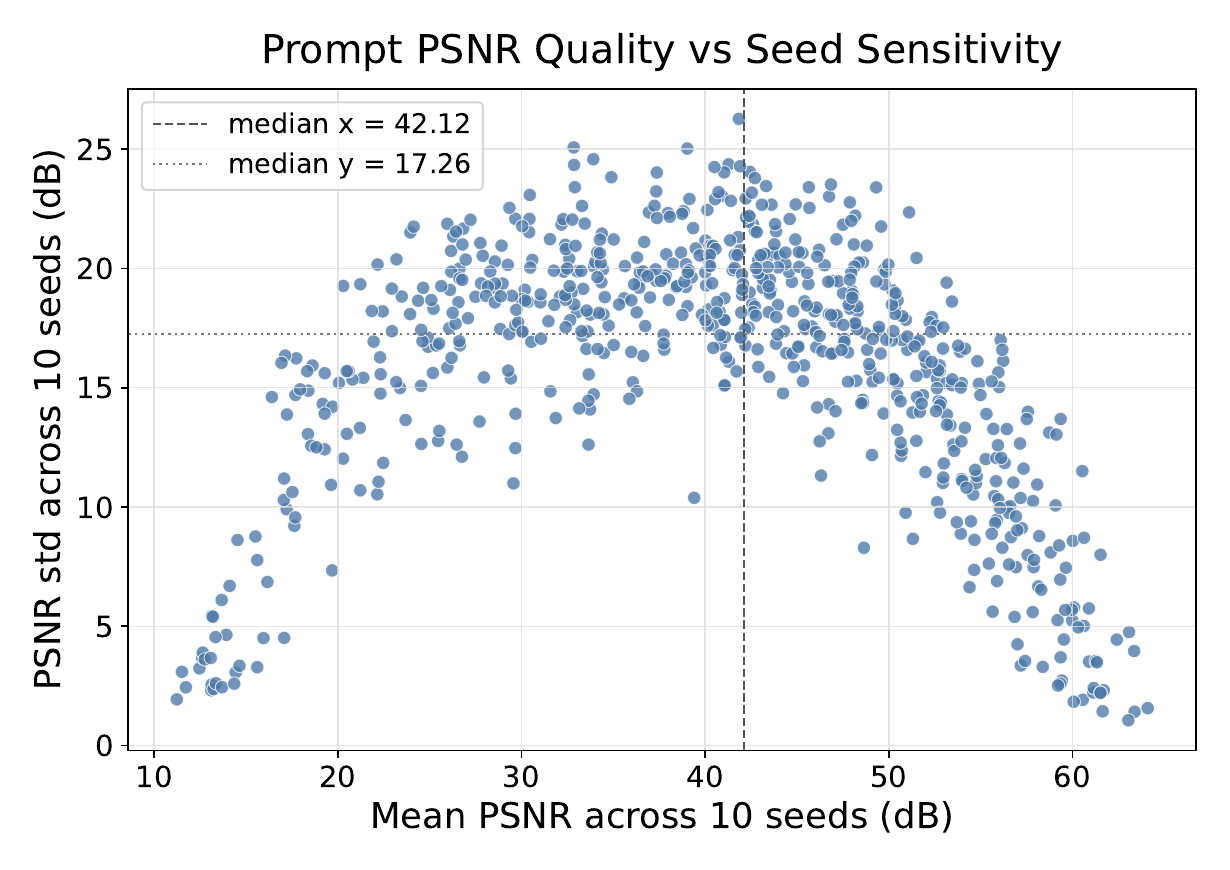}
\caption{
Prompt-level reconstruction behavior across initial latents. Each point
is one prompt, represented by its mean PSNR and standard deviation over
ten seed-indexed initial latents. Prompts with high mean and low
variation are usually reconstructed successfully, prompts with low mean
and low variation usually fail, and prompts with large variation can
succeed or fail depending on the initial latent.
}
\label{fig:prompt_psnr_mean_std}
\end{figure}

Figure~\ref{fig:prompt_psnr_mean_std} reveals a structured pattern in
this two-dimensional summary. Prompts with high mean PSNR tend to have
low standard deviation: they are reconstructed well for most initial
latents. Prompts with low mean PSNR also tend to have low standard
deviation: they fail for most initial latents. In contrast, prompts in
the intermediate mean-PSNR range often have much larger standard
deviation, indicating that reconstruction can succeed for some initial
latents and fail for others.

\begin{figure*}[t]
    \centering
    \includegraphics[width=0.95\linewidth]{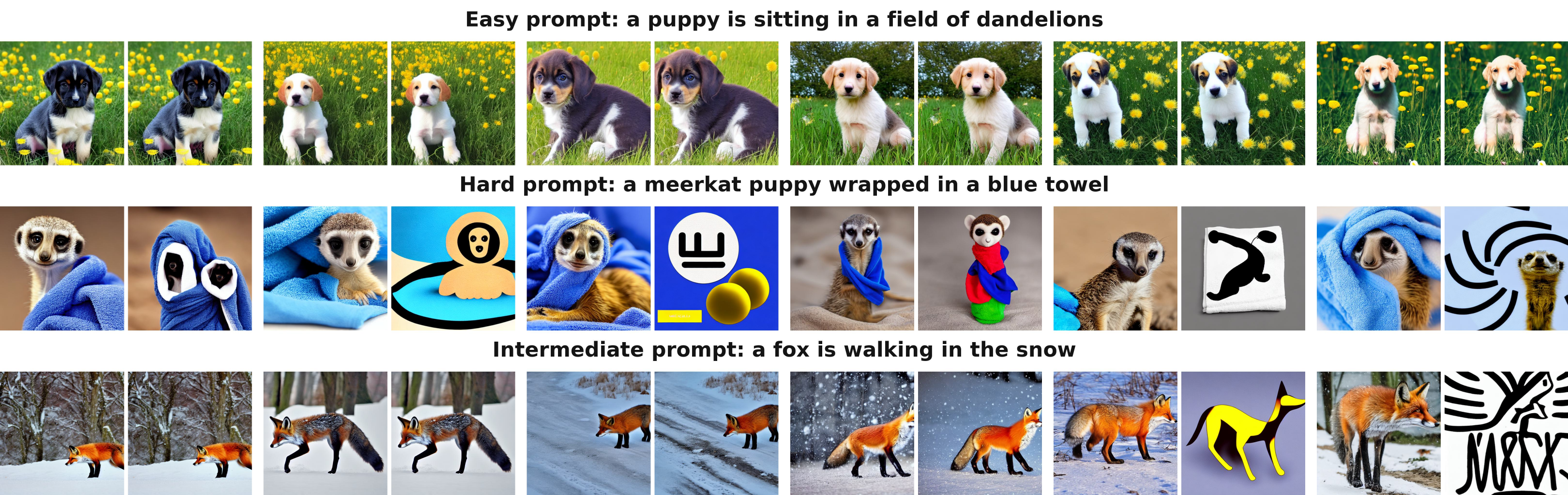}
    \caption{
    Qualitative examples of easy, hard, and intermediate prompts. Each row
fixes one prompt and shows six generated--reconstructed pairs from
different randomly sampled initial latents, with the generated image on the
left and the FPI reconstruction on the right. From top to bottom: easy
prompts reconstruct for most latents, hard prompts fail for most
latents, and intermediate prompts succeed or fail depending on the
initial latent.
    }
    \label{fig:qualitative_montage}
\end{figure*}

We refer to these three regions as easy, hard, and intermediate
prompts, respectively. Figure~\ref{fig:qualitative_montage}
shows representative examples. Easy and hard prompts are relatively insensitive
to the initial latent, but in opposite directions: reconstruction
usually succeeds for the former and fails for the latter. Intermediate
prompts are different. Their reconstruction outcome depends strongly on
the initial latent. More examples are presented in the supplementary material.

As a secondary main-effect check, Table~\ref{tab:aggregate_psnr}
summarizes reconstruction PSNR marginalized over one factor at a
time. Prompt-wise means vary widely, whereas seed-wise means are tightly
concentrated, indicating that none of the ten seed-indexed initial
latents is globally easy or globally hard across prompts. An additive
two-way ANOVA gives the same main-effect trend: the prompt main effect
explains 38.9\% of the variance, while the latent main effect explains
0.56\%. Since each prompt--latent cell has a single observation,
prompt--latent interaction is pooled into the residual; this check
therefore does not explain the latent-sensitive failures observed for
intermediate prompts. Details are provided in the supplementary material.




\begin{table*}[t]
\centering
\caption{
Reconstruction PSNR statistics for the FPI
generation--inversion--reconstruction benchmark. ``All pairs''
summarizes all prompt--latent pairs. ``Prompt-wise mean'' averages over
the ten seed-indexed initial latents for each prompt. ``Latent-wise
mean'' averages over all prompts for each seed-indexed initial latent.
}
\label{tab:aggregate_psnr}
\begin{tabular}{lcccccc}
\toprule
Quantity & $n$ & Mean & Std. & Min & Median & Max \\
\midrule
All pairs & 7000 & 40.61 & 20.35 & 6.08 & 46.11 & 69.61 \\
Prompt-wise mean & 700 & 40.61 & 12.70 & 11.24 & 42.12 & 64.09 \\
Latent-wise mean & 10 & 40.61 & 1.61 & 38.90 & 40.52 & 44.25 \\
\bottomrule
\end{tabular}
\end{table*}

\subsection{Implications}
\label{sec:implications}

The main finding of this section is the three-regime structure of
prompt-level inversion difficulty. Some prompts are easy regardless of
the initial latent, some are hard regardless of the initial latent, and
intermediate prompts are sensitive to the initial latent. The marginal
statistics and ANOVA refine this picture: they show that the ten
initial latents do not act as globally good or bad latents across all
prompts, but they do not explain the prompt-dependent latent sensitivity
observed in the intermediate regime.

This changes the interpretation of inversion failure. Failure is not
well described either as a purely prompt-only effect or as a purely
latent-only effect. Instead, the prompt appears to define the inversion
regime, while the initial latent can determine success or failure within
the intermediate regime. This motivates the next section, where we
analyze generation-side prompt pressure and ask how prompt-conditioned
trajectories differ across these regimes.

\section{Measuring how prompts alter the generation trajectory}
\label{sec:prompt_pressure}

\subsection{Defining prompt pressure}

The previous section shows that reconstruction behavior is not described
by a prompt-only or latent-only explanation. Instead, prompts exhibit
different reconstruction regimes, and intermediate prompts can succeed
or fail depending on the seed-indexed initial latent. Unlike Section~\ref{sec:prompt_seed}, where
the same ten initial latents are shared across all prompts, the analyses
below sample initial latents independently for each prompt and seed. To further analyze
these prompt--latent interations, we examine the generation trajectory itself. Since high
classifier-free guidance often makes inversion more difficult, we ask
whether reconstruction difficulty can be related to how strongly the
prompt-conditioned update deviates from the unconditional denoising
update during generation.

Let \(z_{t-1}\) be a latent produced by the forward DDIM update
\begin{equation}
\label{eq:ddim}
    z_{t-1} =
    k_1(t) z_t + k_2(t) \tilde{\epsilon}_\theta(z_t, t, p; \omega),
\end{equation}
where \(k_1\) and \(k_2\) are timestep-dependent coefficients defined
by the DDIM scheduler. The classifier-free guided prediction is
\begin{equation}
    \tilde{\epsilon}_\theta(z_t, t, p; \omega)
    =
    \epsilon_\theta(z_t,t,\emptyset)
    +
    \omega
    \bigl(
    \epsilon_\theta(z_t,t,p)
    -
    \epsilon_\theta(z_t,t,\emptyset)
    \bigr),
\end{equation}
where \(\emptyset\) denotes the null-text condition. We define the
prompt-induced guidance direction as
\begin{equation}
    \Delta_t(z_t,p)
    =
    \epsilon_\theta(z_t,t,p)
    -
    \epsilon_\theta(z_t,t,\emptyset).
\end{equation}
This term measures the direction by which the prompt changes the
denoising update relative to the unconditional trajectory.

We summarize its contribution to the actual DDIM update by the
step-wise prompt pressure
\begin{equation}
    \Pt(z_t,p)
    =
    \omega |k_2(t)| \|\Delta_t(z_t,p)\|_2 .
\end{equation}
For each prompt--seed pair, we aggregate this quantity over the
generation trajectory:
\begin{equation}
    P^{\mathrm{tot}}_{p,s}
    =
    \sum_t \Pt(z_{t,p,s},p),
\end{equation}
where \(z_{t,p,s}\) denotes the latent at timestep \(t\) generated from
prompt \(p\) and initial latent \(z_{p,s}\). If prompt-induced
guidance makes the trajectory harder to invert, larger
\(P^{\mathrm{tot}}_{p,s}\) should be associated with lower
reconstruction PSNR.

\subsection{Prompt pressure and reconstruction quality}
\label{sec:pressure_global}

\begin{figure*}[t]
\centering
$\begin{array}{cc}
\includegraphics[width=0.45\linewidth]{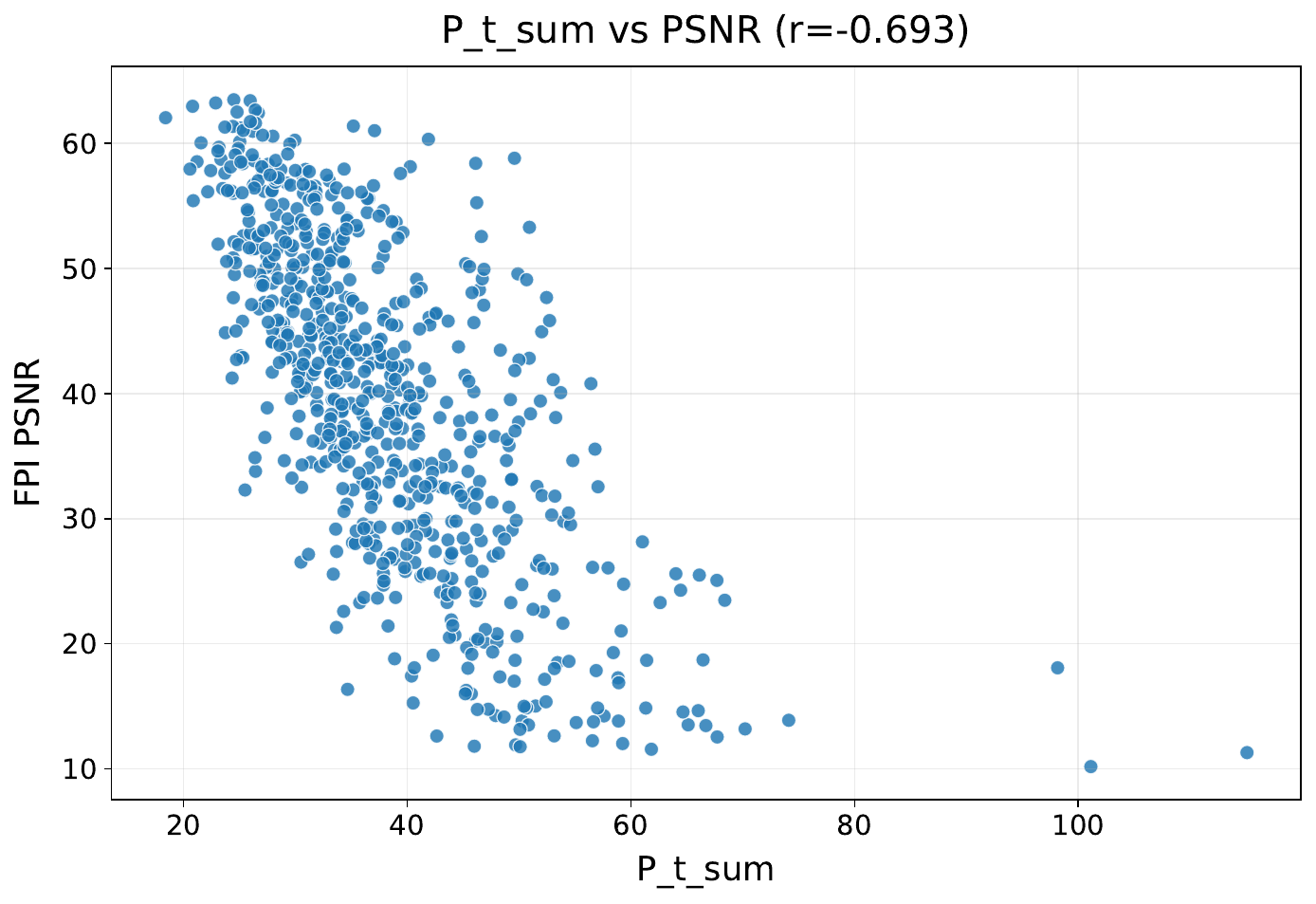}
&
\includegraphics[width=0.45\linewidth]{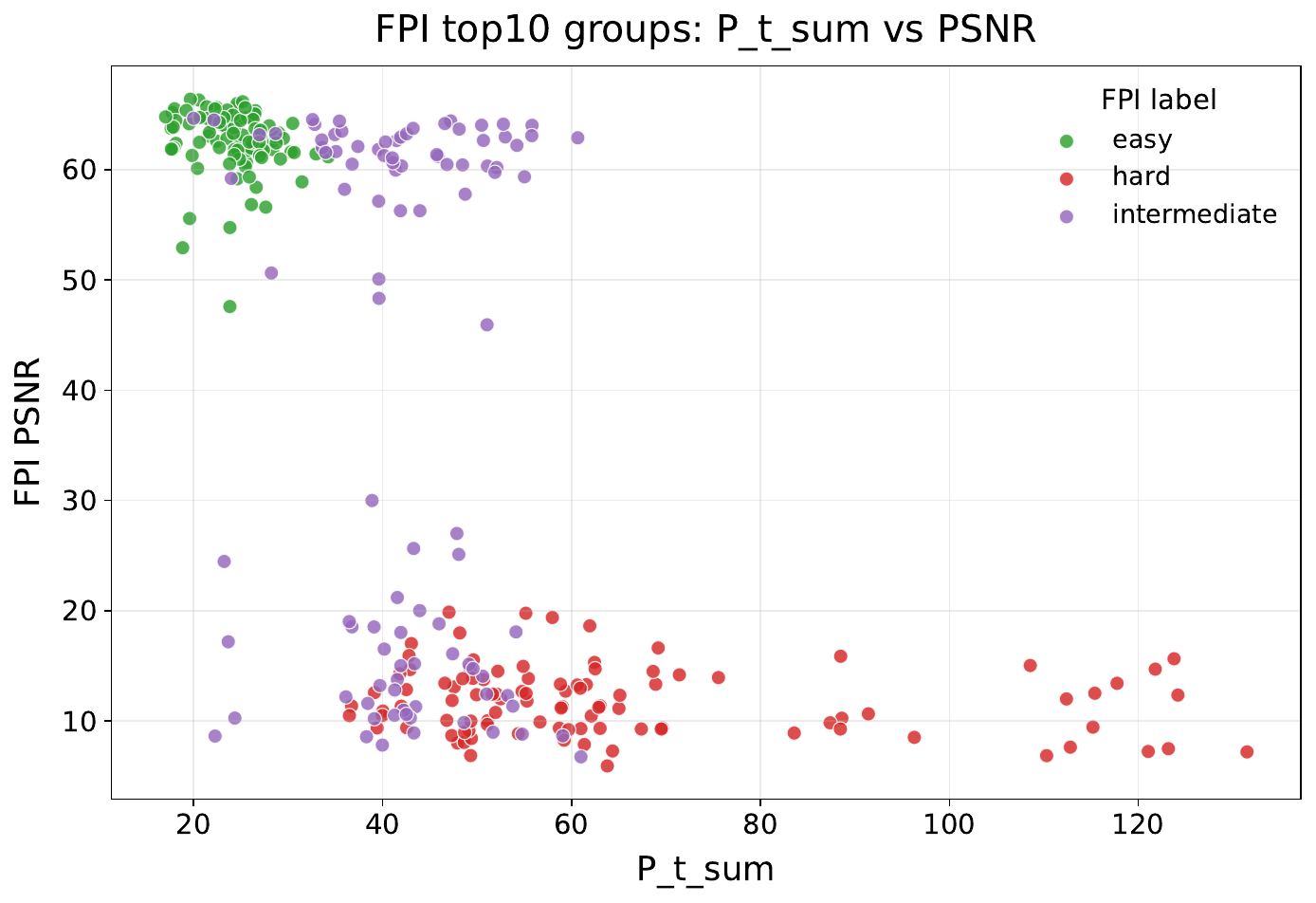}\\
\mbox{\small (a)} & \mbox{\small (b)}
\end{array}$
\caption{
Total prompt pressure versus FPI reconstruction PSNR.
(a) Prompt-level averages over all 700 prompts show a negative overall
correlation between pressure and reconstruction quality.
(b) Prompt--seed samples from the easy, hard, and intermediate groups
show that pressure separates easy and hard prompts, but not the
success/failure split within intermediate prompts.
}
\label{fig:seed_mean_pressure}
\end{figure*}

We first compare total prompt pressure with reconstruction quality over
the full benchmark. Figure~\ref{fig:seed_mean_pressure}(a) plots
seed-averaged total pressure against seed-averaged reconstruction PSNR
for the 700 prompts. The Pearson correlation is \(-0.693\) at the
prompt level and \(-0.458\) at the individual prompt--seed level.

This global trend suggests that \(P^{\mathrm{tot}}\) captures a real
generation-side factor: prompts that push the denoising trajectory more
strongly away from the unconditional update tend to be harder to
reconstruct. However, the relationship is only partial. Prompts with
similar pressure can still have different reconstruction quality, and
pressure alone does not explain the latent-dependent success and failure
observed for intermediate prompts.

\subsection{Pressure for easy, hard, and intermediate prompts}
\label{sec:pressure_groups}

We next examine the three prompt groups identified in
Section~\ref{sec:prompt_regimes}: easy, hard, and intermediate. These
groups are selected from reconstruction statistics only, not from
pressure values. This allows us to test whether generation-side pressure
separates the groups after they have been defined by inversion behavior.

\begin{table*}[t]
\centering
\caption{
Prompt pressure for selected easy, hard, and intermediate prompts. Each
row summarizes 10 prompts and 10 initial latents. Total
pressure separates easy and hard prompts at the group level, but does
not explain the large within-prompt variation of intermediate prompts.
}
\label{tab:selected_pressure}
\begin{tabular}{lrrrr}
\toprule
Group & PSNR mean & PSNR std. & \(P^{\mathrm{tot}}\) mean & \(r(P^{\mathrm{tot}},\mathrm{PSNR})\) \\
\midrule
Easy & 62.62 & 2.89 & 24.28 & -0.145 \\
Hard & 11.79 & 3.04 & 65.00 & -0.156 \\
Intermediate & 40.06 & 23.48 & 42.36 & -0.089 \\
\bottomrule
\end{tabular}
\end{table*}

Table~\ref{tab:selected_pressure} shows a clear group-level ordering.
Easy prompts have low total pressure and high reconstruction PSNR,
whereas hard prompts have much higher total pressure and low PSNR.
Thus, for the stable regimes at both ends of the distribution,
generation pressure provides a plausible coarse explanation: easy
prompts are associated with weak prompt-induced deviations, while hard
prompts are associated with strong deviations.

The intermediate group behaves differently. Its mean pressure lies
between the easy and hard groups, but its PSNR standard deviation is
much larger. Within this group, total pressure is only weakly correlated
with PSNR. This suggests that intermediate failures are not simply
caused by prompt--latent pairs having larger prompt pressure.

Figure~\ref{fig:seed_mean_pressure}(b) visualizes the same point at the
prompt--seed level. Easy and hard prompts occupy different pressure
ranges, but intermediate prompts contain large reconstruction
differences at similar pressure values. For example, the prompt
``a plant against a wall'' has two seeds with almost identical
\(P^{\mathrm{tot}}\) values, 22.26 and 22.14, but their reconstruction
PSNRs are 8.62 dB and 64.50 dB. The total prompt pressure is nearly
unchanged, yet one reconstruction fails and the other is almost exact.
Thus, pressure magnitude does not resolve the success/failure split
inside the intermediate regime.

\subsection{Summary of pressure analysis}
\label{sec:pressure_summary}

The pressure analysis gives a mixed picture. At the global level,
larger prompt-induced guidance is associated with lower reconstruction
quality. Across the selected prompt groups, total pressure also
separates the stable regimes: easy prompts have low pressure, while
hard prompts have high pressure. Therefore, \(P^{\mathrm{tot}}\) is a
useful coarse descriptor of prompt-level inversion difficulty.

However, pressure does not explain the key latent-sensitive cases.
Intermediate prompts can succeed or fail depending on the initial
latent, but these outcomes are not reliably separated by total prompt
pressure. We also inspected the temporal profile of \(\Pt\) over the
50 generation steps, including normalized pressure curves, peak values,
entropy, and Gini concentration. These statistics did not reveal a
consistent pattern that separates successful and failed intermediate
cases. In particular, samples with similar total pressure and similar
pressure profiles can still have very different reconstruction PSNR.

Thus, prompt pressure captures part of the generation-side structure
behind inversion difficulty, but it is not a complete predictor. The
remaining failures in the intermediate regime likely depend on more
local trajectory properties than the scalar magnitude or simple temporal
shape of prompt-induced guidance.

\section{Prompt content and wording}
\label{sec:prompt_components}

The previous sections show that prompt information matters for
reconstruction difficulty, even though it does not fully explain the
success or failure of individual intermediate prompt--latent pairs.
Here we set aside this fine-grained intermediate-regime question and ask
a simpler one: what distinguishes prompts that are usually reconstructed
well from prompts that are not? We approach this question from the text
side by perturbing prompt content and wording.

A motivating contrast comes from two prompts with similar surface
structure: ``a squirrel is sitting on top of a wooden fence'' and
``a yellow apple sitting on top of a wooden table.'' The former has a
mean FPI reconstruction PSNR of 14.23 dB, placing it in a hard regime,
whereas the latter has a mean PSNR of 44.63 dB. Since both prompts use
a similar ``sitting on top of'' construction, this contrast suggests
that the object and scene content, not only the syntactic template, may
affect whether a prompt is easy to invert.

\subsection{Varying the main object and scene context}
\label{sec:modifier_grid}

We first decompose simple prompts into a subject and a context. The
subject denotes the main visual object, such as ``squirrel'' or
``apple,'' while the context denotes the remaining scene phrase, such as
``wooden fence'' or ``wooden table.'' For example, in ``a squirrel is
sitting on top of a wooden fence,'' the subject is ``squirrel'' and the
context is ``wooden fence.'' In ``a yellow apple sitting on top of a
wooden table,'' the subject is ``apple'' and the context is ``wooden
table.''

Using the PIE-Bench prompt pool as a starting point, we construct a
controlled subject--context grid with 14 subjects and 10 contexts.
The subjects are
\begin{quote}
\small
squirrel, bee, rat, owl, fox, rabbit, duck, apple, mushroom,
kitten, cat, puppy, bird, and dog.
\end{quote}
The contexts are
\begin{quote}
\small
transparent glass jar, wooden fence, snowy field, streetlight night,
wooden table, bare tree branch, rock, shallow water, and grass.
\end{quote}

We evaluate every subject--context combination with randomly sampled
initial latents, resulting in 140 prompts and 1400 prompt--seed pairs.

\begin{figure*}[t]
\centering
\includegraphics[width=0.9\linewidth]{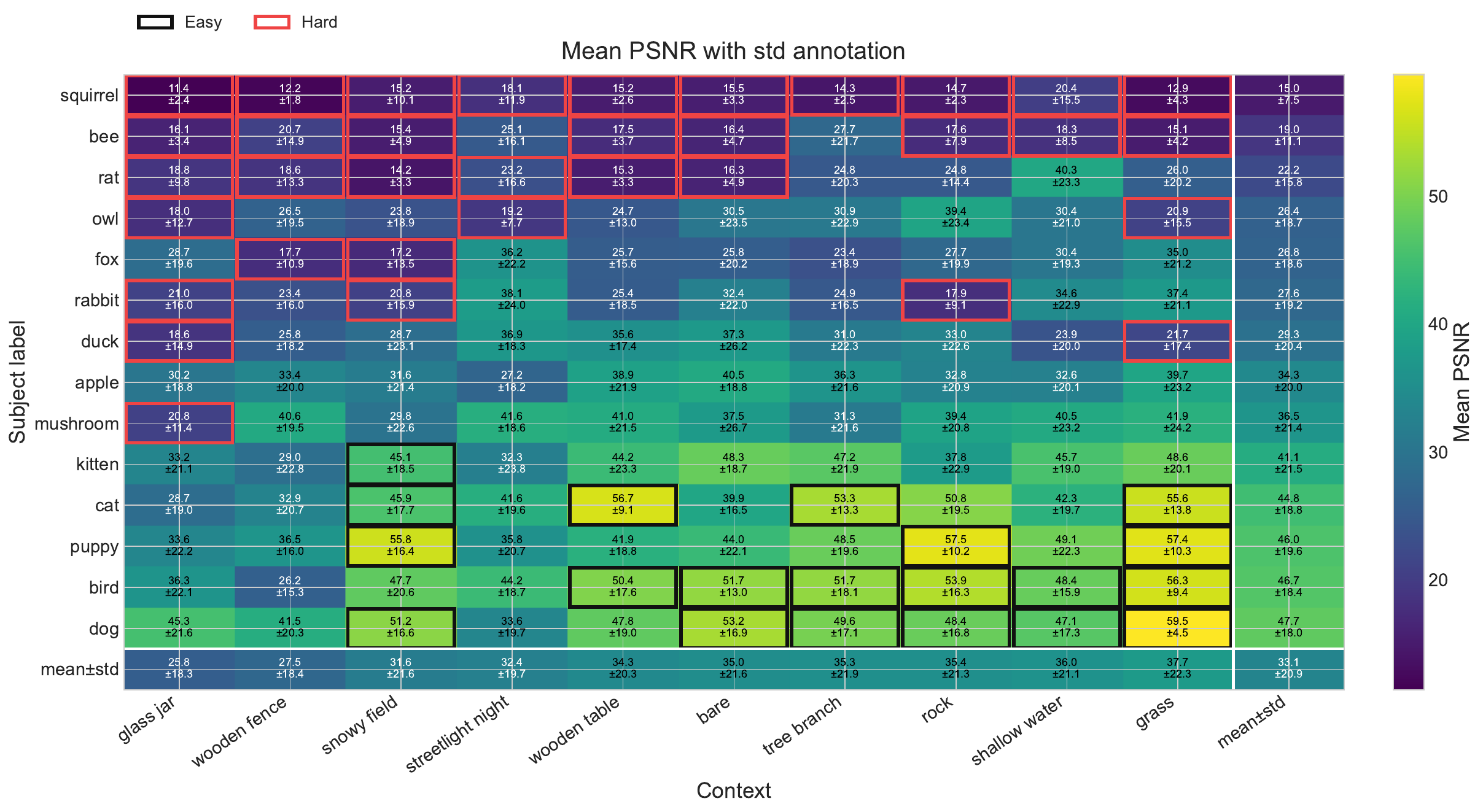}
\caption{
Effect of changing the main object and scene context on FPI
reconstruction PSNR. Rows vary the main object and columns vary the
context. Each cell reports mean PSNR over ten randomly sampled initial
latents. Reconstruction quality changes more strongly with the main
object than with the surrounding context.
}
\label{fig:modifier_grid}
\end{figure*}

Figure~\ref{fig:modifier_grid} shows the mean reconstruction PSNR for
each cell. The strongest visible structure is along the subject axis.
For example, squirrel remains hard across contexts, with a subject-level
mean of 14.99 dB, while dog, bird, puppy, and cat are much easier, with
subject-level means of 47.70, 46.68, 46.01, and 44.76 dB, respectively.
Context also changes the reconstruction quality, but its effect is
weaker: context-level means range from 25.76 dB for transparent glass
jar to 37.71 dB for grass. Qualitative examples are provided in the supplementary material.

A simple additive two-factor decomposition gives the same descriptive
trend. Subject identity explains 25.3\% of the total PSNR variance,
whereas context identity explains 3.0\%; the remaining 71.6\% is
residual and subject--context interaction. We use this decomposition
only as a coarse summary, but it supports the view that the main visual
concept is one axis separating easy prompts from prompts that are more
difficult to reconstruct.

\subsection{Changing wording while preserving meaning}
\label{sec:paraphrase_grid}

We next ask whether the exact wording of a prompt changes
reconstruction difficulty when the intended concept is preserved. We
construct paraphrase grids for three base concepts: a cat sitting on a
wooden table, a squirrel sitting on a wooden fence, and the Great Wave
off Kanagawa. For each base concept, we generate ten paraphrases using
ChatGPT and evaluate each paraphrase with the same ten seed-indexed
initial latents. 
The paraphrases include lexical replacement, e.g., changing ``a cat is
sitting on a wooden table'' to ``a cat seated on the surface of a wooden
table''; sentence restructuring, e.g., changing ``a squirrel is sitting
on top of a wooden fence'' to ``a wooden fence with a squirrel sitting
on it''; and more specific descriptions, e.g., changing ``The Great
Wave off Kanagawa'' to ``the traditional ukiyo-e piece The Great Wave
off Kanagawa.'' Detailed paraphrase grids and reconstruction result examples are provided in the supplementary material.

\begin{figure}[t]
\centering
\includegraphics[width=0.9\linewidth]{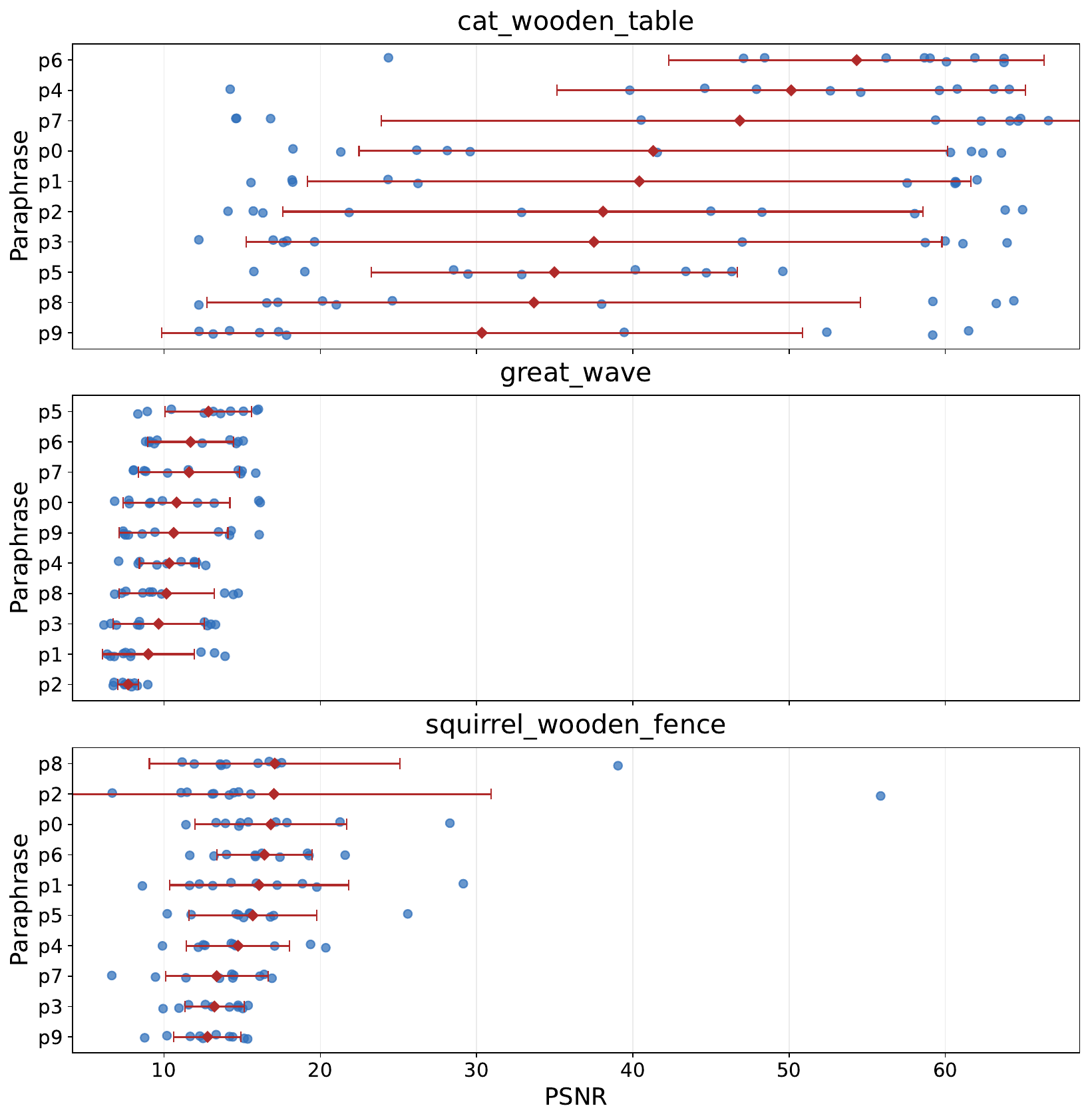}
\caption{
Effect of wording changes on FPI reconstruction PSNR. Each panel shows
ten paraphrases of the same base concept, evaluated over ten
randomly sampled initial latents. Blue points are individual latents; red
diamonds and error bars show the mean and standard deviation for each
paraphrase. Paraphrases are sorted by mean PSNR.
}
\label{fig:paraphrase_effect}
\end{figure}

Figure~\ref{fig:paraphrase_effect} shows different behavior across the
three concepts. The cat/table concept is sensitive to wording:
paraphrase means span 30.34--54.32 dB. For example, ``a cat seated on
the surface of a wooden table'' reaches 54.32 dB, while ``a cat
positioned on top of a brown wooden table'' drops to 30.34 dB. Thus,
wording can move a borderline concept between easier and more difficult
reconstruction regimes.

In contrast, the squirrel/fence and Great Wave concepts remain hard
under paraphrasing. Their paraphrase-level means span only
12.78--17.08 dB and 7.70--12.84 dB, respectively. In these cases,
changing the wording does not move the prompt out of the hard regime.
This suggests that some hard prompts are tied to the underlying visual
concept rather than to a particular surface phrasing.


Overall, these prompt-side experiments do not explain the seed-specific
success and failure of intermediate prompts. They instead identify a
coarser effect: prompt content, especially the main visual object, can
shift prompts toward easy or non-easy reconstruction regimes, while
surface wording can further modulate borderline cases.


\section{Testing guidance consistency during inversion}
\label{sec:diagnostic_intervention}

The preceding analysis suggests that inversion failure is not explained
by a single global prompt-strength statistic. In particular,
intermediate prompts can contain both successful and failed
reconstructions at similar total prompt pressure. This motivates a
limited diagnostic question: if high-CFG failures arise from local
instability along the inversion trajectory, can reconstruction be
improved by responding to that instability locally, rather than by
using a globally low guidance scale?

This section is not intended to introduce a new general-purpose
inversion method or a comprehensive benchmark. Instead, we test this
question with a compact intervention on top of fixed-point inversion.
The intervention keeps the inversion trajectory at the target high
guidance scale whenever the fixed-point solve is stable, and lowers the
guidance scale only when the current timestep fails to converge. We
refer to this diagnostic intervention as \textsc{SkipInv}.

\subsection{Reducing guidance only at unstable steps}
\label{sec:local_guidance_relaxation}

At each inverse timestep, given \(z_{t-1}\), fixed-point inversion
solves the inverse DDIM step as an implicit fixed-point problem:
\begin{equation}
z_t = g_t(z_t; z_{t-1}, p, \omega),
\end{equation}
where \(g_t\) denotes the one-step inverse update induced by the DDIM
sampler. We use a relaxed fixed-point update,
\begin{equation}
z_t^{(i+1)}
=
(1-\alpha_i)z_t^{(i)}
+
\alpha_i g_t(z_t^{(i)}; z_{t-1}, p, \omega),
\end{equation}
with residual
\begin{equation}
\ell_i
=
\left\|
g_t(z_t^{(i)}; z_{t-1}, p, \omega)
-
z_t^{(i)}
\right\|_2^2 .
\end{equation}
When the residual stops decreasing, we first damp the update by reducing
\(\alpha_i\). If the timestep still fails to reach a fixed convergence
threshold, we lower the guidance scale and retry the same timestep.
Once the guidance scale has been lowered, the reduced value is used for
later inverse steps, giving a monotone non-increasing inversion
schedule
\begin{equation}
\omega_1 \geq \omega_2 \geq \cdots \geq \omega_T .
\end{equation}
During reconstruction or editing, we replay the denoising process with
the recorded schedule in the forward order.

This intervention does not add editing controls, optimize text
embeddings, or modify attention maps. It only changes the fixed-point
iteration and the per-step guidance schedule. We therefore use it as a
diagnostic test of trajectory consistency, not as a replacement for
existing inversion frameworks. Implementation details and hyperparameter
settings are given in the supplementary material.

\subsection{Reconstruction and editing results}
\label{sec:trajectory_consistency_results}

We compare only the settings needed to test guidance consistency. As a
strong iterative inversion baseline, we use AIDI~\cite{pan2023aidi}.
AIDI-GS\(a\)/GS\(b\) denotes inversion with guidance scale \(a\),
followed by reconstruction or editing with guidance scale \(b\). Thus,
AIDI-GS1/GS7 represents the common low-guidance-inversion /
high-guidance-reconstruction setting, while AIDI-GS7/GS7 matches the
guidance scale in both stages. \textsc{SkipInv} starts from the matched
high-guidance setting and relaxes guidance only at locally unstable
inverse steps.

For reconstruction, we report two latent-space scores using
\(-\log(\mathrm{MSE})\): Init--Inv, the error between the true initial
latent and the inverted latent, and Gen--Rec, the error between the
generated latent and the reconstructed latent. Higher is better for
both metrics. For editing, we use Prompt-to-Prompt~\cite{hertz2022prompt}
and report CLIP text score for target-prompt alignment and PSNR between
the edited and source images for source preservation.

\begin{table*}[t]
\centering
\caption{
Reconstruction and editing results for matched and mismatched guidance
settings. Reconstruction uses latent \(-\log(\mathrm{MSE})\), and
editing uses CLIP text score and source-preservation PSNR. The upper
bound edits from the true initial latent and is not an inversion method.
}
\label{tab:trajectory_consistency}
\begin{tabular}{l|rr|rr}
\toprule
Method & Init--Inv & Gen--Rec & CLIP text & Edit PSNR \\
\midrule
AIDI-GS1/GS7 & 4.03 & 1.82 & 0.3081 & 17.56 \\
AIDI-GS7/GS7 & 9.05 & 9.83 & 0.3079 & 20.92 \\
\textsc{SkipInv} & 9.35 & 12.21 & 0.3102 & 22.30 \\
Upper bound & -- & -- & 0.3121 & 22.76 \\
\bottomrule
\end{tabular}
\end{table*}

Table~\ref{tab:trajectory_consistency} supports the trajectory-consistency
hypothesis. The mismatched setting, AIDI-GS1/GS7, gives poor latent
reconstruction: Init--Inv is 4.03 and Gen--Rec is 1.82. Matching the
guidance scale in AIDI-GS7/GS7 substantially improves both scores to
9.05 and 9.83. This indicates that low-guidance inversion is not merely
a harmless stabilization trick; it can produce a latent that is poorly
aligned with the high-guidance reconstruction dynamics.

\textsc{SkipInv} further improves the matched high-guidance setting,
reaching 9.35 for Init--Inv and 12.21 for Gen--Rec. The editing results
show the same qualitative pattern. AIDI-GS1/GS7 obtains reasonable text
alignment but low source preservation, while AIDI-GS7/GS7 improves Edit
PSNR from 17.56 dB to 20.92 dB. \textsc{SkipInv} improves the tradeoff
again, reaching a CLIP text score of 0.3102 and an Edit PSNR of
22.30 dB, close to the upper-bound setting that edits from the true
initial latent.

These results are not intended as a broad method benchmark. Instead,
they isolate one practical implication of the analysis: when the target
generation or editing process uses high CFG, inversion benefits from
remaining close to the high-guidance trajectory and relaxing guidance
only at locally unstable steps.

\section{Conclusion}
\label{sec:conclusion}

This paper analyzed high-\(\cfg\) diffusion inversion in a controlled
generation--inversion--reconstruction setting, where the true initial
latent and denoising trajectory are known. Our results show that
inversion difficulty is not explained by a purely prompt-only or
latent-only view. Instead, prompts form three types of prompt-level reconstruction behavior:
easy prompts that succeed for most seed-indexed initial latents, hard
prompts that fail for most latents, and intermediate prompts whose
success depends on the prompt--latent pairing.

Generation-side prompt pressure provides a useful but incomplete
explanation. It is negatively correlated with reconstruction quality and
separates easy from hard prompts at a coarse level, but neither its
total magnitude nor its simple temporal profile explains the
success/failure split within intermediate prompts. Text-side analyses
show that the main visual subject and wording can change inversion
difficulty, while still leaving latent-dependent failures unresolved.

As a diagnostic check, we also evaluate a compact trajectory-consistency
intervention. The results suggest that high-\(\cfg\) inversion benefits
from remaining close to the target high-guidance trajectory and relaxing
guidance only at locally unstable steps. Overall, our findings point to
a local, trajectory-dependent view of inversion failure, shaped jointly
by prompt content and the initial latent rather than by a single global
prompt-strength or seed-quality factor.

{\small
\bibliographystyle{ieee_fullname}
\bibliography{egbib}
}

\end{document}